\pdfoutput=1
\documentclass[journal,twoside,web]{ieeecolor}
\usepackage{tmi}
\usepackage{cite}
\usepackage{amsmath,amssymb,amsfonts}
\usepackage{algorithmic}
\usepackage{graphicx}
\usepackage{textcomp}
\usepackage{caption}
\usepackage{xurl}
\usepackage{booktabs}
\usepackage{multirow}
\usepackage{mathtools}
\usepackage{txfonts}
\usepackage{soul}
\usepackage{placeins}
\usepackage{textcmds}
\usepackage[switch]{lineno}

\makeatletter
\let\NAT@parse\undefined
\makeatother
\usepackage{hyperref} 

\title{High-Resolution Maps of Left Atrial Displacements and Strains Estimated with 3D Cine MRI using Online Learning Neural Networks}

\author{Christoforos Galazis, Samuel Shepperd, Emma Brouwer, Sandro Queirós, Ebraham Alskaf, Mustafa Anjari, Amedeo Chiribiri, Jack Lee, Anil A. Bharath, and Marta Varela
\thanks{C. Galazis is with the Department of Computing, Imperial College London, London, UK (email: c.galazis20@imperial.ac.uk}
\thanks{S. Shepperd and E. Brouwer were with the Department of Physics, Imperial College London, London, UK. E Brouwer is currently with the Spinoza Centre for Neuroimaging, Amsterdam, The Netherlands (email: samuel.shepperd19@imperial.ac.uk and brouwer@herseninstituut.knaw.nl)}
\thanks{S. Queirós is with the Life \& Health Sciences Research Institute (ICVS), School of Medicine, University of Minho, Braga, Portugal, and the ICVS/3B’s - PT Government Associate Laboratory, Braga/Guimarães, Portugal (email: sandroqueiros@med.uminho.pt)}
\thanks{E. Alskaf, A. Chiribiri and J. Lee are with the School of Biomedical Engineering \& Imaging Sciences, King's College London, London, UK (email: ebraham.alskaf@kcl.ac.uk, amedeo.chiribiri@kcl.ac.uk and jack.lee@kcl.ac.uk)}
\thanks{M. Anjari is with the Department of Radiology, Royal Free Hospital, London, UK and  the Department of Brain Repair \& Rehabilitation, UCL Queen Square Institute of Neurology, London, UK (email: m.anjari@ucl.ac.uk)}
\thanks{A. A. Bharath is with the Department of Bioengineering, Imperial College London, London, UK (email: a.bharath@imperial.ac.uk)}
\thanks{M. Varela is with the National Heart \& Lung Institute, Imperial College London, London, UK (email: marta.varela@imperial.ac.uk)}
}

\let\oldtwocolumn\twocolumn
\renewcommand\twocolumn[1][]{%
    \oldtwocolumn[{#1}{
    \begin{center}
           \vspace{-40pt}
           \textbf{Aladdin: Left Atrium Motion Analysis Workflow}\par\medskip
           \includegraphics[width=1.0\textwidth]{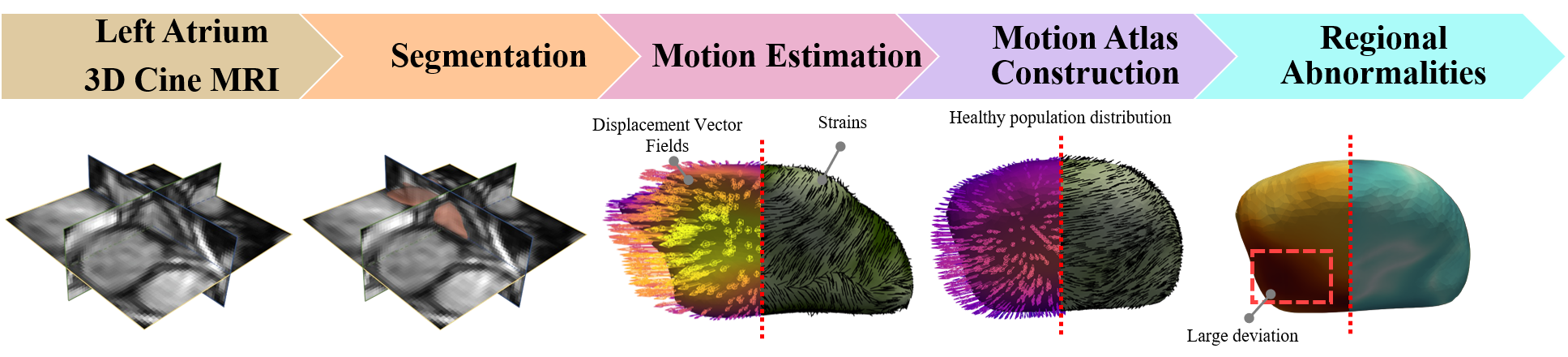}
           \captionof{figure}{Workflow of the Analysis for Left Atrial Displacements and Deformations using onlIne learning neural Networks, \textit{Aladdin}: \textbf{(A)} Take as input a 3D high-resolution sequence of MR images of the left atrium (LA) and segmentation maps of 3 cardiac phases. \textbf{(B)} Predict segmentation of all cardiac phases using a segmentation network. \textbf{(C)} Predict the Displacement Vector Field (DVF) of the LA image contours across the cardiac cycle using an image registration network and calculate the surface principal strains. \textbf{(E)} Construct a displacement and strain LA atlas combining data from 10 healthy subjects. \textbf{(F)} Identify regional motion abnormalities by estimating the Mahalanobis distance between a given patient's data and the atlas distribution.}
           \label{fig:overview}
        \end{center}
        
    }]
    
}

\begin{document}
\bstctlcite{IEEEexample:BSTcontrol}

\maketitle

\begin{abstract}

%%%% Abstract
The functional analysis of the left atrium (LA) is important for evaluating cardiac health and understanding diseases like atrial fibrillation. Cine MRI is ideally placed for the detailed 3D characterization of LA motion and deformation but is lacking appropriate acquisition and analysis tools. Here, we propose tools for the Analysis for Left Atrial Displacements and DeformatIons using online learning neural Networks (Aladdin) and present a technical feasibility study on how Aladdin can characterize 3D LA function globally and regionally. Aladdin includes an online segmentation and image registration network, and a strain calculation pipeline tailored to the LA. We create maps of LA Displacement Vector Field (DVF) magnitude and LA principal strain values from images of 10 healthy volunteers and 8 patients with cardiovascular disease (CVD), of which 2 had large left ventricular ejection fraction (LVEF) impairment. We additionally create an atlas of these biomarkers using the data from the healthy volunteers. Results showed that Aladdin can accurately track the LA wall across the cardiac cycle and characterize its motion and deformation. Global LA function markers assessed with Aladdin agree well with estimates from 2D Cine MRI. A more marked active contraction phase was observed in the healthy cohort, while the CVD $\text{LVEF}_\downarrow$ group showed overall reduced LA function. Aladdin is uniquely able to identify LA regions with abnormal deformation metrics that may indicate focal pathology. We expect Aladdin to have important clinical applications as it can non-invasively characterize atrial pathophysiology. All source code and data are available at: \url{https://github.com/cgalaz01/aladdin_cmr_la}.

\end{abstract}


\section{Introduction}
\label{sec:introduction}
%\subsubsection{Left Atrium}
%\IEEEPARstart{T}{he} motion and deformation of the left atrium (LA) are impaired in conditions such as heart failure (HF) \cite{hoit2017evaluation,peters2021left}.

\IEEEPARstart{D}{isturbances} in atrial function are increasingly recognized as significant contributors to the development and progression of various cardiovascular diseases (CVDs), such as atrial fibrillation, atrial myopathy, heart failure, and stroke \cite{smiseth2022imaging,hoit2017evaluation, blume2011left,inoue2022echocardiographic,shen2019atrial}. These atrial functional abnormalities are typically part of disease-induced remodelling processes \cite{shen2019atrial,bisbal2020atrial}. They can be accompanied by tissue-level changes, especially fibrotic remodelling or fatty deposition, and also organ-level remodelling, manifesting as changes in atrial size and shape. Organ-level remodelling is expected to be a later manifestation of CVD than changes at the cellular or tissue-level. 

Current assessments of left atrial (LA) motion and deformation typically only capture global or one-dimensional changes in the atria. LA strains are usually estimated from single-plane long-axis echocardiograms \cite{montserrat2015left} or magnetic resonance imaging (MRI) \cite{peters2021left}. This is in contrast to the left ventricle (LV), for which several dedicated 3D regional metrics of motion and deformation exist \cite{nabeshima2020review}, with known clinical correlates \cite{xu2017value,sugano2017value,hayat2012comparison,medvedofsky20182d}. This discrepancy highlights a lack of tools for a detailed 3D assessment of LA function and deformation. Here, we address this gap, introducing a novel end-to-end acquisition and image processing tool, Aladdin, that enables comprehensive 3D analysis of LA motion and deformation. We also present proof-of-concept analysis of LA function in 10 healthy volunteers and 8 CVD patients, to demonstrate Aladdin's future clinical potential.

\subsection{LA Strains}
\label{lab:strains}
The LA plays a vital role in cardiac function by modulating left ventricular filling and overall cardiac output. It operates through three phases: 1) reservoir phase, in which the LA is filled with oxygenated blood from the pulmonary veins, reaching its maximum volume; 2) conduit phase, where blood is passively emptied into the LV, reducing the LA volume; and 3) booster pump, in which the LA actively contracts, further emptying the chamber to its minimum volume \cite{hoit2014left}. Markers of LA motion and deformation, called \qq{LA strains}, are typically assessed across these phases using 2D cardiac long-axis through echocardiograms or dynamic (Cine) MRI \cite{badano2018standardization,smiseth2022imaging}.

The global LA strain is a single value that describes the change in LA perimeter in a 4-chamber view \cite{badano2018standardization}. It is the simplest and mostly reported strain value \cite{badano2018standardization}. In 2D regional strain analysis, the change in length of specific segments of the LA contour is measured \cite{badano2018standardization}. These LA strains, and their temporal derivatives, \qq{LA strain rates}, have been correlated with atrial fibrillation (AF) burden, heart failure (HF) burden, presence of LA fibrosis, heart valve disease and cardiomyopathies \cite{hoit2017evaluation,montserrat2015left,thomas2020evaluation}.

While these 2D LA strains markers are valuable, they have several limitations. As they are estimated from a single view, they suffer from sampling bias and characterize only 2D changes. Moreover, they provide an incomplete characterization of tissue deformations compared to the definition of strain in the physical sciences. In engineering, strains are tensors, which quantify the complex directional material deformations (3D stretches and shears) of a body in response to applied stresses. 3D full coverage strain measurements of the LA have the potential to capture vital information concerning tissue biomechanics, thereby expanding clinical applications \cite{peters2021left}. These could include the identification of atrial regions of abnormally reduced strain, which could be associated with stiffer, fibrotic LA myocardium. Current LA fibrosis identification relies on late gadolinium-enhanced (LGE) MRI, a very subjective contrast agent technique with poor reproducibility \cite{chubb2018reproducibility,fahmy2021improved}. Furthermore, the use of LGE MRI carries the risk of gadolinium deposition, which can have adverse effects and may be contraindicated in certain patients \cite{fraum2017gadolinium}. Thus, there is a great need for better imaging tools for \textit{in vivo} non-invasive LA tissue characterization.

As in the LV, regional strains may contribute to fibrosis identification \cite{lota2021prognostic}. Regional strain metrics are also likely to be early disease markers, preceding changes in global functional biomarkers such as ejection fraction \cite{duchateau2020machine}. Thus, high-resolution spatio-temporal 3D maps of LA deformation can potentially provide more specific, earlier signatures of CVD pathology that affects the atria, with enhanced diagnostic and prognostic value \cite{fraum2017gadolinium}.

\subsection{Cine MRI}
Cine MRI is increasingly used to characterize LA motion, given its excellent signal-to-noise ratio, spatial resolution and operator-independence when compared to speckle tracking echocardiography (STE)\cite{schuster2016cardiovascular,voigt20192}. Cine MRI, which is widely included in clinical cardiac MRI examinations, allows the reconstruction of MR images at various cardiac phases, enabling a dynamic evaluation of cardiac motion. The LA is visible, although incompletely, in the 2-chamber, 4-chamber, and 3-chamber views, each acquired in a separate breath-hold as thick slices. These images have been used to study LA motion and deformation \cite{smiseth2022imaging,habibi2016cardiac,schuster2019left}, usually analyzed using feature tracking (FT) techniques \cite{smiseth2022imaging}. FT is a post-processing step \cite{schuster2016cardiovascular} which involves the identification and tracking across the cardiac cycle of key features (e.g., manually or automatically delineated LA wall regions). These deformation analyses are inherently incomplete, as they are conducted as thick single-slice images that lack full LA coverage. Moreover, at the typical spatial resolution used for ventricular 2D Cine MR imaging (slice thickness: 8-10$~mm$, in-plane resolution: $\sim 2~mm$ \cite{kramer2020standardized}), partial volume effects preclude an accurate identification and motion characterization of the LA wall. Different slices are also acquired at different breath-holds, further complicating 3D motion analysis. Nonetheless, 2D FT has been found clinically useful for the LA \cite{xu2022state}. Previous studies have analysed LA function using stacks of thick slices covering the LA acquired in a short-axis view \cite{lourencco2021left,uslu2021sa} but 3D FT has not yet been explored due to the lack of 3D Cine MRI protocols for the LA.

\subsection{Segmentation}

Delineating (segmenting) the LA is critical for assessing its structure and function, particularly given its complex anatomy. Manual segmentation of the LA in 2D long-axis and short-axis Cine MRI demonstrates high agreement, with inter-observer intra-class correlation coefficient (ICC) ranging from 0.81 to 0.99 and intra-observer ICC ranging from 0.87 to 0.98 \cite{alfuhied2021reproducibility,gonzales2021automated,tondi2023use,zareian2015left}. Similarly, 3D LGE-MRI also shows strong observer agreement \cite{li2022medical}, with inter-observer ICC between 0.79 and 0.97 and intra-observer ICC between 0.95 and 0.98 \cite{muargulescu2019reproducibility}.

Manual segmentation of medical images, which involves partitioning an image into anatomically meaningful regions, is an extremely repetitive and time-consuming process. Improved reproducibility and ease-of-use can be achieved through automated segmentation algorithms \cite{gonzales2021automated}, which are predominately deep neural network (NN)-based. In cardiac MRI, U-Net \cite{ronneberger2015u} and its variants have been successfully deployed for LA segmentation of LGE-MRI \cite{uslu2021net,xiong2021global,kausar20233d,wong2022gcw,li2022atrialjsqnet}. More recently, nnU-Net \cite{isensee2021nnu} builds on the U-Net architecture and introduces an automated training pipeline including preprocessing and postprocessing steps, leading to impressive results on a variety of medical images, including cardiac MRI.

\subsection{Displacement Vector Field}
A necessary step for characterizing atrial deformations is the tracking of the position of regions of interest in the images across the cardiac cycle. From these, displacement vector fields (DVFs), the vectors that link material points in the myocardium across different phases of the cardiac cycle, can be estimated.

DVFs are usually computed by aligning (registering) images across the cardiac cycle, as we propose here, or by using optical flow techniques such as block matching \cite{schuster2016cardiovascular}. Strain tensors can be estimated through mathematical manipulation of the DVFs. 

NNs are increasingly used for image registration, offering advantages such as improved accuracy and scalability over traditional techniques, but they also face challenges like the need for large labeled datasets and potential issues with generalizability \cite{haskins2020deep,fu2020deep,o2020deep}. Recently, various NN-based image registration methods have been proposed to analyze LV deformations across the cardiac cycle from Cine MR images. Common architectures are based on: U-Net \cite{morales2021deepstrain,sinclair2022atlas,zhang2023learning,voxelmorph}, variational autoencoders \cite{bello2019deep,qin2023generative,krebs2019learning}, or Siamese NN \cite{yu2020foal,qin2018joint}. Usually, the displacement field is learned in an unsupervised way by using a spatial resampling module \cite{jaderberg2015spatial} that estimates the DVFs between pairs of NN-registered images.

% U-Net {upendra2020convolutional}

Segmentation maps of cardiac structures have been used with cardiac motion tracking networks to improve their performance. Typically, this can be accomplished by incorporating the segmentation outputs as part of the input \cite{bello2019deep}, constraining the DVFs so the transformed segmentation maps agree with the target segmentation \cite{morales2021deepstrain}, or integrating them into a joint segmentation-motion tracking task \cite{qin2018joint,sinclair2022atlas}.

% DVFs so the transformed segmentations agree with the target segmentation \cite{zheng2019explainable,morales2021deepstrain}

Out-of-the-box techniques designed for the LV are therefore unlikely to succeed in the LA, as the LA noticeably differs from the LV. The LA has much thinner myocardial walls \cite{whiteman2019anatomical,varela2017atrial,whitaker2016role}, a more irregular and complex morphology \cite{whiteman2019anatomical,ho2012left}, and high structural diversity between individuals \cite{whiteman2019anatomical}. We therefore propose a specific registration NN for the LA as part of the Aladdin tool.

\subsection{Atlas}
The clinical value of LA regional motion characterization hinges on the ability to identify regional LA motion abnormalities. For this, LA maps of reference values of regional LA deformation biomarkers (such as DVF magnitude or principal strain values) are needed. These LA maps can be constructed by building an LA atlas of motion biomarkers. In this context, an atlas is a statistical model that characterizes the features of a specific population \cite{young2009computational}.

An atlas is typically created using registration techniques to align multiple representations (e.g., segmentation maps) of a given organ across a specific population \cite{ericsson2008construction,sinclair2022atlas,bai2015bi}. Regional biomarkers (such as LA displacements and strains) can be mapped onto the atlas common reference space, to enable direct statistical and visual analyses across subjects \cite{ericsson2008construction,bai2015bi}. Atlases serve as standardized references for analyzing and comparing imaging data, and can provide a method for comprehensive characterization of normal LA motion and deformation patterns \cite{young2009computational,medrano2013atlas,bai2015bi}. Furthermore, they allow for the identification of regional differences in biomarkers between pathological cases and the healthy population \cite{gilbert2020artificial}.

% deformation patterns \cite{young2009computational,rodero2021linking,medrano2013atlas,bai2015bi}

\subsection{Overview and Aims}
% Our hypothesis is that the utilization of 3D high-resolution strain maps can improve the functional analysis of the LA, thereby improving the prognostic and diagnostic capabilities available for cardiac conditions.
% To achieve such high spatial resolution across the cardiac cycle, 3D Cine MRI is required. However, there are currently no such publicly available MRI datasets or adequate image analysis tools to extract the DVF and strain maps of the whole LA using MRI. This is primarily due to a lack of recognition of the potential benefits of tailoring MR techniques to image the LA at high spatial resolution. 

%Leveraging 3D high-resolution strain maps can prove to be beneficial for the functional analysis of the LA, thereby improving the prognostic and diagnostic capabilities available to us. To achieve such high spatial resolution across the cardiac cycle, 3D Cine MRI is required. However, there are currently no publicly available MRI datasets or adequate image analysis tools to extract DVF and strain maps of the whole LA using MRI. This is due to a lack of MR techniques tailored to image the LA at high spatial resolution.\hl{Is it a lack of techniques or that the value is not seen in implementing these techniques?}

We propose \textit{Aladdin}, Analysis of Left Atrial Displacements and DeformatIons with neural Networks, to automatically characterize LA displacements and strains from high-resolution 3D Cine MRI. We conduct a technical feasibility study to characterize LA displacements and strains in small healthy and CVD patient cohorts. We also create the first atlas of regional LA displacements and deformations across the cardiac cycle. As shown in Figure \ref{fig:overview}, Aladdin includes:
\begin{enumerate}
    \item An online learning segmentation NN, nnU-Net \cite{isensee2021nnu}, for the semi-automatic segmentation of the LA across the cardiac cycle.
    \item An online weakly supervised learning image registration NN, Aladdin-R, to automatically estimate LA DVFs across the cardiac cycle.
    \item An algorithm to calculate regional LA strains across the cardiac cycle, which treats the LA myocardium as an infinitesimally thin 2D surface.
    \item The construction of a LA DVF and strain atlas using data from 10 healthy volunteers.
    \item Proof-of-principle characterization of regional LA strains of cardiovascular patients using the created atlas.
\end{enumerate}

\section{Methods}
The source code for Aladdin, the atlas, as well as the anonymized Cine MR images, are publicly available: \url{https://github.com/cgalaz01/aladdin_cmr_la} and \url{https://zenodo.org/records/13645121}.

\subsection{LA Cine MR Images}
We use 3D Cine MRI balanced Steady State Free Precession (bSSFP) scans of the LA acquired using a novel high-resolution acquisition protocol \cite{varela2020strain}. The images are electrocardiogram (ECG)-gated and acquired in a single 25-second breath-hold using a 1.5T Philips Ingenia MRI scanner with a 32-channel cardiac coil. The typical field of view used is $400 \times 270 \times 70~mm^3$ and the acquisition matrix is $256 \times 256 \times 36$ to give a reconstructed resolution of $1.72 \times 1.72 \times 2.00~mm^3$. We use SENSE with factors of 2.3 and 2.6 along each phase encode direction, and 55\% view sharing for a total of 20 phases across the cardiac cycle. Cardiac phase 0 corresponds to the ECG R-peak, representing ventricular end-diastole. This corresponds to atrial end-systole at which the LA is expected to have the smallest volume. An example cross-section of the 3D Cine images can be seen in Supplementary Figure (SupFig) \href{https://github.com/cgalaz01/aladdin_cmr_la/blob/main/supplements/README.md#SupFig1}{1} and slices across the cardiac cycle from two cases in SupVid \href{https://github.com/cgalaz01/aladdin_cmr_la/blob/main/supplements/README.md#SupVid1}{1} and \href{https://github.com/cgalaz01/aladdin_cmr_la/blob/main/supplements/README.md#SupVid2}{2}.

We analyse images acquired after informed consent and under ethical approval from 18 subjects: 10 healthy volunteers (age: $30.4 \pm 4.7$ years; female: $40.0\%$), and 8 patients with cardiovascular disease (CVD) (age: $53.8 \pm 15.7$ years; female $62.5\%$). As this dataset was acquired to evaluate the clinical feasibility of the new LA protocol \cite{varela2020strain}, patients do not necessarily have abnormal LA atrial function. The pathologies present in the CVD group are: myocarditis (1 patient), history of syncope (2), myocardial infarction (3), non-ischaemic cardiomyopathy (1), and hypertrophic cardiomyopathy (1). Two of these patients had a reduced LV ejection fraction ($\leq$40\% \cite{esc2016}). Specifically, one with severe abnormality (22\%) with confirmed diagnosis of myocardial infarction and the other with moderate abnormality (35\%) with a diagnosis of hypertrophic cardiomyopathy and myocardial infarction. The remaining cases have either normal ($\geq$50\%) or mild abnormality (40-49\%). We therefore divide these two patients into a subgroup of \qq{CVD $\text{LVEF}_\downarrow$} for further comparison.

To provide training and benchmarking data for Aladdin, the LA was manually segmented across the entire cardiac cycle for all subjects by two experts. The pulmonary veins (PV) and LA appendage (LAA) were excluded from the segmentation maps, as we focus on functional rather than structural variations in the atlas.

%\begin{figure}[!ht]
%\centering
%  \textbf{High Resolution 3D Cine MRI}\par\medskip
%  \includegraphics[width=0.35\textwidth]{figures/methods/data_example.png}
%  \caption{\textbf{Top:} Example cross section of the LA from a healthy volunteer imaged using the dedicated 3D high resolution acquisition protocol. \textbf{Bottom:} LA volume across the cardiac cycle from the same healthy volunteer. Phases 0, 8 and 15, in which the LA was manually segmented (see Section \ref{sec:preprocessing}), correspond approximately to the transitions between the reservoir, conduit and booster pump phases.}
%  \label{fig:la_data_example}
%\end{figure}

\subsection{Preprocessing}
\label{sec:preprocessing}
The LA images are preprocessed to standardize their characteristics before being inputted to the NNs: they are cropped to a size of $96 \times 96 \times 36$ voxels, centred at the LA, and their intensity is min-max normalized between $[0,1]$. We also perform a rigid-body translation of the LA to ensure that its approximate centroid, estimated from the predicted segmentation maps, is stationary across the cardiac cycle.

%Our tool utilizes LA segmentation maps from cardiac phases 0, 8 and 15, which are manually segmented in all subjects. The pulmonary veins and appendage are excluded from the segmentation maps to avoid identifying structural rather than functional variabilities on the atlas. 

%Moreover, 5 healthy volunteers and 5 CVD subjects have their LA manually segmented across the whole cardiac cycle by an expert. These additional segmentation maps across the cardiac cycle are used as ground truth data for evaluating the segmentation and image registration models.

\begin{figure*}[!ht]
\centering
  \includegraphics[width=1.0\textwidth]{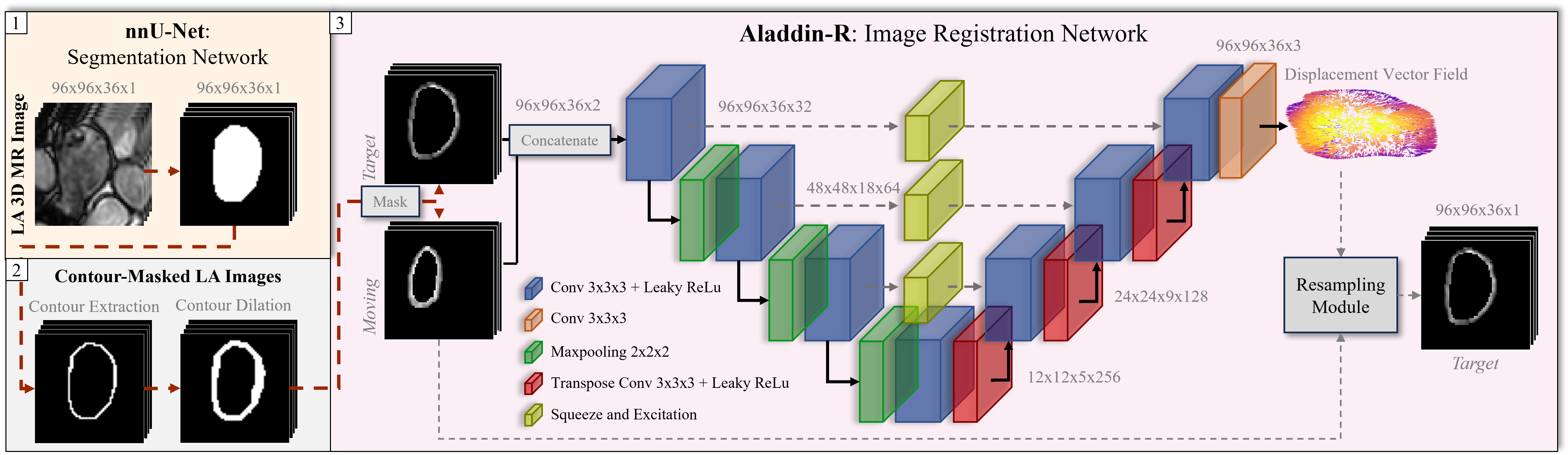}
  \caption{Aladdin's image registration workflow. It consists of: \textbf{1)} an online learning (subject-by-subject basis) nnU-Net that estimates the LA segmentation maps, \textbf{2)} contours from the LA segmentation maps are first extracted and then dilated to create masks for the images, and \textbf{3)} an online learning 3D U-Net-like image registration network, Aladdin-R.
  }
  \label{fig:la_architecture}
\end{figure*}

%Overview of the proposed common architecture of Aladdin-S (\textbf{top}) and Aladdin-R (\textbf{bottom}) and their respective data inputs (\textbf{left}). In total, we have 3 downsampling (decoding) and upsampling (encoding) steps. Each encoding block consists of a $3 \times 3 \times 3$ convolutional layer followed by a $2 \times 2 \times 2$ maxpooling layer. The decoding block consists of a concatenation layer of the output of the previous layer and the respective skip connection, followed by a $3 \times 3 \times 3$ transposed convolutional layer and a $3 \times 3 \times 3$ convolutional layer. After each convolutional layer, a Leaky Rectified Linear Unit (ReLU) activation \cite{maas2013rectifier} with an \textit{alpha} value of $0.3$ is applied to all except the final convolution, which utilizes a sigmoid activation. Glorot uniform initialization \cite{glorot2010understanding} is used to set the initial weights. Both networks are trained independently on a subject-by-subject basis.

\subsection{nnU-Net: Segmentation Neural Network}
\label{sec:segmentation}
To avoid manually segmenting the LA across all phases of the cardiac cycle, we use the state-of-the-art NN for medical image segmentation: nnU-Net \cite{isensee2021nnu}. Our implementation of the nnU-Net takes as input a 3D LA image in a given cardiac phase and predicts a corresponding LA binary segmentation map. We use the manual segmentation maps at the three cardiac phases (0, 8 and 15) as ground truth data for training. Given the limited availability of data, we adopt an online learning approach inspired by \cite{yu2020foal} and \cite{lei2020blind}. In this method, the model's weights are optimized solely using few unseen data from each new subject, starting from a randomly initialized state. This subject-by-subject training ensures that the model uniquely adapts to each individual. Subsequently, using the three manually segmented maps, we can predict the remaining LA segmentation maps for the subject.

\subsection{Aladdin-R: Image Registration Neural Network}

Aladdin-R is a 3D convolutional NN based on the U-Net architecture \cite{ronneberger2015u}, implemented in TensorFlow \cite{tensorflow2015-whitepaper}. As with U-Net, Aladdin-R uses encoding and decoding blocks to capture features at different spatial resolutions, and skip connections to preserve higher-resolution features. Rather than the standard skip connections between each encoder and decoder block we use squeeze and excitation (SE) blocks \cite{hu2018squeeze}, which offer dynamic feature map-wise re-calibration, with a squeeze ratio of 8. The complete architecture can be seen in Figure \ref{fig:la_architecture}.

As we are only interested in the motion of the LA wall, we use masked MR images of the LA myocardium (Figure \ref{fig:la_architecture} left panel) as inputs to the registration NN, Aladdin-R. This approach removes any irrelevant tissue to reduce tracking errors while leaving enough context for the network to distinguish between the LA wall and surrounding tissues. We first extract the contours of the LA segmentation maps yielded by nnU-Net across the cardiac cycle. The LA contours are then dilated using a spherical structure with a radius of 1 voxel. Using this dilated LA contour segmentation, the original LA images are masked out and used as inputs to Aladdin-R.

Aladdin-R takes in a pair of contour-masked 3D LA images from the same patient: one at phase 0 (moving image), and the other at another cardiac phase (target image). The outputs of the model are the deformed moving image and the associated DVFs that encode this non-linear transformation.

For Aladdin-R, the 3D input pairs are combined into a unified 3D image, each represented as a different channel. Aladdin-R's final convolutional layer uses a linear activation function and outputs each vector component of the DVF as a distinct feature map. This DVF is enforced through the usage of a 3D spatial resampling module \cite{jaderberg2015spatial}, similarly to VoxelMorph \cite{voxelmorph}, as a final layer. The resampling module linearly resamples the moving image to the target image given the predicted DVF.

We employ online learning to train Aladdin-R on all the cardiac phases from a single subject for $500$ epochs. We use the Adam optimizer algorithm with parameters specified from \cite{adam_optimizer}. The weights are updated by minimizing the mutual information loss \cite{hoffmann2021synthmorph} (the negative of the mutual information) between the target and the resampled moving image with an empirically set bin size of 128. The learning rate is set to $0.001$, a learning decay rate of $0.8$, and a batch size of $1$.

The image registration task is ill-posed because the image similarity loss does not guarantee a unique optimal DVF solution. Moreover, the expected deformations of the LA across the cardiac cycle should be spatio-temporally smooth to preserve its physical integrity. For these reasons, we add a bending energy regularization term to the loss function \cite{rueckert1999nonrigid}, with an empirically selected regularization weight of $0.1$. This regularizer will guide Aladdin-R towards a physiologically plausible solution.

%We assess Aladdin-R's ability to track the non-dilated LA contour across the cardiac cycle using the HD and DS metrics. These are evaluated between the GT LA contour and LA contour for the same cardiac phase outputted by Aladdin-R. These analyses are performed in the 10 subjects with GT segmentation maps across all phases of the cardiac cycle.

%We also compared the performance of Aladdin-R with state-of-the-art image registration solutions proposed for similar tasks: image registration package Advanced Normalization Tools (ANTs) \cite{avants2009advanced}; and image registration NNs VoxelMorph ($Vxm$) \cite{balakrishnan2018unsupervised} and VoxelMorph constrained on the image segmentation ($Vxm\text- seg$) \cite{voxelmorph}. While the Medical Image Tracking Toolbox (MITT) \cite{mitt} was used in a similar study \cite{varela2020strain}, it is not applicable for contour images.

\subsection{LA Strain Calculation}
Due to the LA's thinness relative to its other dimensions \cite{varela2017novel}, we treat it as an infinitesimally thin surface \cite{kiendl2015isogeometric}. Strains on each LA segment can therefore be represented by a 2D symmetric tensor with two principal strain directions (embedded on the surface LA mesh) and two corresponding principal strain values.

We first obtain a closed surface triangular mesh of the LA by applying the marching cubes algorithm to the LA segmentation from the first cardiac phase using scikit-image \cite{scikit-image}. We then use the PyVista library \cite{sullivan2019pyvista} for the mesh smoothing. 

For each triangular cell in the LA mesh, we calculate the finite Green-Lagrangian strain tensors \cite{mcveigh2001imaging} using cardiac phase 0 as the reference (undeformed) state. We take the following steps:
\begin{enumerate}
    \item We calculate the metric tensor, $g$, of the vertex points in curvilinear coordinates. For each LA mesh triangle with vertex positions $x_0$, $x_1$ and $x_2$, defined in a Cartesian basis, the normalized covariant base $v = [g_{1}, g_{2}, g_{3}]$ can be calculated using:
    \begin{equation}
    \begin{aligned}
    \begin{split}
        & g_{1} = \frac{x_{1} - x_{0}}{||x_{1} - x_{0}||} \\
        & g_{2} = \frac{x_{2} - x_{0}}{||x_{2} - x_{0}||} \\
        & g_{3} = g_{1} \times g_{2} \\
        & g = v^T \cdot v
    \end{split}
    \end{aligned}
    \end{equation}
    
    \item The metric tensors at the reference configuration (cardiac phase 0), $G_{ij}$, and in each of the deformed configurations, $g_{ij}$, can be used to estimate the Green-Lagrange strain tensor $E_{ij}$ \cite[Chapter~5]{le2017nonlinear}:
    \begin{equation}
    \begin{aligned}
        & E_{ij} = \frac{1}{2} (g_{ij} - G_{ij})
    \end{aligned}
    \end{equation}

    \item $E_{ij}$ are represented in local covariant coordinates, which are difficult to interpret. We now rewrite them in Cartesian coordinates for the reference configuration, $E_{kl}$, represented by the orthonormal basis $(e_1, e_2, e_3)$:
    \begin{equation}
    \begin{aligned}
    \begin{split}
        & e_{1} = g_{1} \\
        & e_{2} = \frac{g_{2} - g_{2} \cdot e_{1}}{||g_{2} - g_{2} \cdot e_{1}||} \\
        & e_{3} = \frac{e_{1} \times e_{2}}{||e_{1} \times e_{2}||} \\
    \end{split}
    \end{aligned}
    \end{equation}
    
    \item The Green-Lagrangian strain tensor is calculated as:
     \begin{equation}
     \label{alg:strains}
        E_{kl} = E_{ij} (e_{k} \cdot G^{i})(G^{j} \cdot e_{l}) 
    \end{equation}
    \item We compute the eigenvalues ($\lambda$) and eigenvectors ($v$) of $E$ defined in step \ref{alg:strains}. As the LA is represented by a 2D surface, the strain eigenvalue corresponding to the direction orthogonal to the local LA surface is always 0. The principal strain value and direction of $E$ at each triangular cell in the LA are treated as the regional main biomarkers of LA function.
    
\end{enumerate}

\subsection{LA Atlas Generation} 
\label{sec:methods_atlas}

%\begin{figure*}[!ht]
%\centering
%  \textbf{Atlas Generation}\par\medskip
%  \includegraphics[width=0.5\textwidth]{figures/methods/atlas_pipeline.png}
%  \caption{Construction steps of the average atlas from the LA segmentation maps of the 10 healthy volunteers. It consists of a forward step (green full line) that applies the transformation $T_i$ that registers all cases to the initial reference space ($LA_0$, chosen randomly from the healthy volunteers pool). The forward transformations are then inverted and averaged, to give the average inverse transformed $\bar T^{-1}$. The atlas is created by applying $T_i \bar T^{-1}$ to each case to bring it to the final atlas space.}
%  \label{fig:atlas_pipeline}
%\end{figure*}

We create an atlas of the LA with the segmentation maps from the 10 healthy volunteers, to allow the direct comparison and statistical analyses of the calculated DVF and strain metrics. To construct the atlas, we follow a similar strategy to the brain atlas creation steps detailed in \cite{ericsson2008construction} using the SimpleITK \cite{lowekamp2013design} package.

The first step consists of registering the LA segmentation in cardiac phase 0 of each case ($i$) to an arbitrary selected reference case, in a forward transformation ($T_i$) step. This consists of successive affine and non-rigid diffeomorphic registrations. 

Then, to avoid biasing the atlas to the randomly selected reference case, we calculate the average inverse transformation ($\overline{T}^{-1}$): $\overline{T}^{-1} = \frac{1}{n}\sum_{i=1}^{n}{T_{i}^{-1}}$. This has proven to be a robust strategy in minimizing potential biases of the atlas towards the initial reference case \cite{ericsson2008construction}.
%\cite{ericsson2008construction,kuklisova2011dynamic,sanchez2012age}.

By applying $T\overline{T}^{-1}$ to each $i$, we transform each case into the atlas space. We overlay these transformed cases to obtain the atlas consensus segmentation map which requires at least 50\% agreement between all cases. Then, we obtain the mesh of the atlas segmentation map using marching cubes. Additionally, we apply $T\overline{T}^{-1}$ to the left and right PV, and LAA landmark points of each case. The average positions of these landmarks are used to indicate their locations on the atlas.

The final step consists of the registration of the DVF and strain fields estimated in each subject to the atlas space, using appropriate change of basis transforms. 

We construct the final atlas by obtaining the mean, standard deviation and coefficients of variation (CV) of the motion and deformation metrics: magnitude of DVFs, principal strain directions and principal strain values. These are calculated at each LA atlas vertex, over all healthy subjects and across all cardiac phases.

As a potential tool to detect focal abnormalities in the LA, we can evaluate differences in regional LA function in a given subject in comparison to the healthy subjects' atlas. For this, we register that case directly to the atlas using the affine and non-rigid methods employed for atlas creation. Then, to quantify individual deviations from the atlas we use the Mahalanobis Distance (MD) \cite{de2000mahalanobis}:
\begin{equation}
MD_{vt}(\mathbf{s_{vt}}) = \sqrt{(\mathbf{s_{vt}} - \boldsymbol{\mu_{vt}})^T \Sigma^{-1}_{vt} (\mathbf{s_{vt}} - \boldsymbol{\mu_{vt}})}
\end{equation}
where $\mathbf{s}$ is the subject's metric, $\boldsymbol{\mu}$ is the atlas mean of the metric, $\boldsymbol{\Sigma}$ the atlas covariance, $\mathbf{v}$ indexes the vertex, $\mathbf{t}$ the cardiac phase.

\subsection{Experimental Setup}
We evaluate the models using the Hausdorff distance (HD, in $mm$) and Dice score (DS) metrics. HD measures the maximum distance between nearest points on the ground truth (GT) and predicted segmentation maps, with lower values indicating better performance. DS measures overlap, with 1 indicating perfect overlap. These analyses are performed across all phases of the cardiac cycle for each subject. We also measured inter-observer variability in 15 cases (7 healthy, 8 with CVD) and intra-observer variability in 6 cases (3 healthy, 3 with CVD) using the Dice score. Models were trained with three different random seeds, and results are presented as the average and standard deviation.

We compared Aladdin-R with state-of-the-art image registration methods: Advanced Normalization Tools (ANTs) \cite{avants2009advanced}, VoxelMorph (Vxm) \cite{balakrishnan2018unsupervised}, and VoxelMorph constrained on LA segmentation maps (Vxm-seg) \cite{voxelmorph}. The Medical Image Tracking Toolbox (MITT) \cite{mitt}, used in a similar study \cite{varela2020strain}, is not applicable for contour images.

We also compared the estimated 3D strains from Aladdin-R with the global longitudinal strains (GLS) derived from the 2-chamber and 4-chamber views. The 2D strains are calculated as the Green-Lagrangian GLS, which represents the relative change in length of the LA perimeter \cite{buggey2018left}. This perimeter is measured by tracing the endocardial border of the LA throughout the cardiac cycle, excluding the mitral valve annulus, LAA, and PVs \cite{lang2015recommendations}.

Furthermore, we evaluated the estimated LA ejection fraction (LAEF), defined as $(max\_volume - min\_volume) / max\_volume \times 100\%$, and LA active ejection fraction (LAaEF), defined as $(preactivation\_volume - min\_volume) / preactivation\_volume \times 100\%$, between the 3D and long-axis views.

The approximate LA volume from the 2D long-axis view is estimated using the formula \cite{lang2005recommendations}: $\frac{8}{3 \times \pi} \times \frac{A^2}{L}$, where $A$ represents the area of the LA in either the 2-chamber or 4-chamber view, and $L$ is the length from the midpoint of the mitral annular plane to the back wall of the corresponding chamber. For improved estimation, the combined 2-chamber and 4-chamber and the formula is updated to \cite{lang2005recommendations}: $\frac{8}{3 \times \pi} \times \frac{A_{2ch} \times A_{4ch}}{\min(L_{2ch}, L_{4ch})}$, where $A_{2ch}$ and $A_{4ch}$ are the areas of the 2-chamber and 4-chamber views, respectively, and $\min(L_{2ch}, L_{4ch})$ is the minimum length between the two views.

All networks were trained on a single Nvidia RTX 6000. For training and inference of one case, nnU-Net and Aladdin-R required approximately 1 hour, and 20 minutes, respectively. ANTs and the atlas construction and registration were executed on a 3XS Intel Core i7 (10700K, 3.8GHz, 8 Core). To construct the atlas on the 10 healthy cases required 2.5 hours, while registering an individual case to the atlas required 6 minutes.

\section{Results}
Our proposed method successfully estimated high-resolution LA displacement and strain maps across the cardiac cycle from all 18 subjects. We additionally created an atlas of LA motion and deformation biomarkers in healthy volunteers, and showed how it can help identify preliminary differences in these biomarkers in the CVD patients.

\subsection{LA Segmentation}

\begin{figure}[!ht]
\centering
  \includegraphics[width=0.5\textwidth]{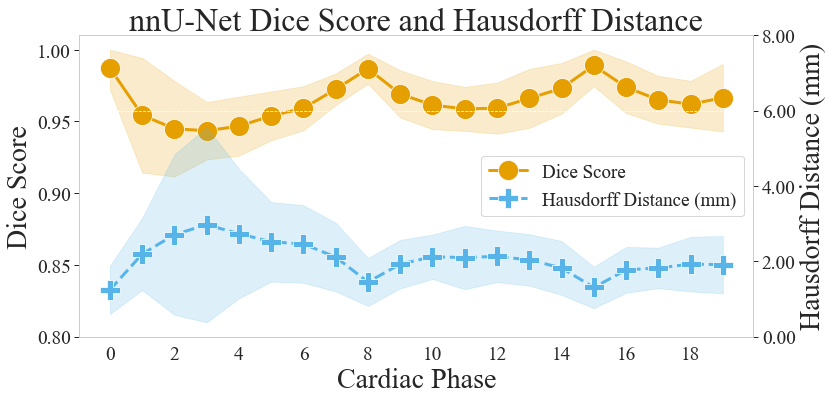}
  \caption{The spatially averaged Dice score (orange line) and Hausdorff distance (mm) (blue dotted line) for the online learning nnU-Net across the cardiac cycle when trained on cardiac phases 0, 8, and 15 across all 18 subjects. The shaded areas around each line, colored to match the respective mean, indicate the standard deviation.}
  \label{fig:la_seg_results}
\end{figure}

The online learning nnU-Net, with a total of 16.1M trainable parameters, demonstrates high accuracy in segmenting the LA throughout the cardiac cycle, despite learning from augmented segmentation maps from 3 cardiac phases only. In the 10 healthy cases, we achieve excellent HD of $2.42 \pm 0.81~mm$ and DS of $0.97 \pm 0.01$. We observe similarly good results on the 8 CVD cases with a HD of $2.13 \pm 1.15~mm$ and a DS of $0.97 \pm 0.02$. These values compare very well with the measured intra-observer and inter-observer agreement: DS of $0.96 \pm 0.06$ and $0.79 \pm 0.16$, respectively. 

Although nnU-Net performs better on cardiac phases that are closer to the manual segmentation maps used for training (see Figure \ref{fig:la_seg_results}), our data augmentation scheme successfully prevents overfitting to the phases it was trained on. Representative segmentation results can be seen in Supplementary Videos (SupVid) \href{https://github.com/cgalaz01/aladdin_cmr_la/blob/main/supplements/README.md#SupVid3}{3} and \href{https://github.com/cgalaz01/aladdin_cmr_la/blob/main/supplements/README.md#SupVid4}{4}.

\subsection{LA Registration}
Aladdin-R accurately tracks the LA wall across the cardiac cycle, including in cardiac phases with large displacements (see SupVid \href{https://github.com/cgalaz01/aladdin_cmr_la/blob/main/supplements/README.md#SupVid5}{5}). When comparing the segmentation of the registered LA with the ground truth LA segmentation maps for the same cardiac phase, we obtain excellent HD of $1.14 \pm 0.32~mm$ and DS of $0.98 \pm 0.02$. 

The estimated DVF from Aladdin-R preserves the LA's expected rapid volume increase during the reservoir phase (0-8), followed by passive and active emptying during the conduit (9-15) and booster pump phases respectively (16-19). Aladdin-R consistently outperforms all other evaluated registration techniques in accuracy (see Table \ref{tab:registration_results}) and spatio-temporal smoothness (see SupVid \href{https://github.com/cgalaz01/aladdin_cmr_la/blob/main/supplements/README.md#SupVid6}{6} and \href{https://github.com/cgalaz01/aladdin_cmr_la/blob/main/supplements/README.md#SupVid7}{7}).

\begin{figure}[!ht]
\centering
  \textbf{Healthy and CVD Comparison}\par\medskip
  \includegraphics[width=0.45\textwidth]{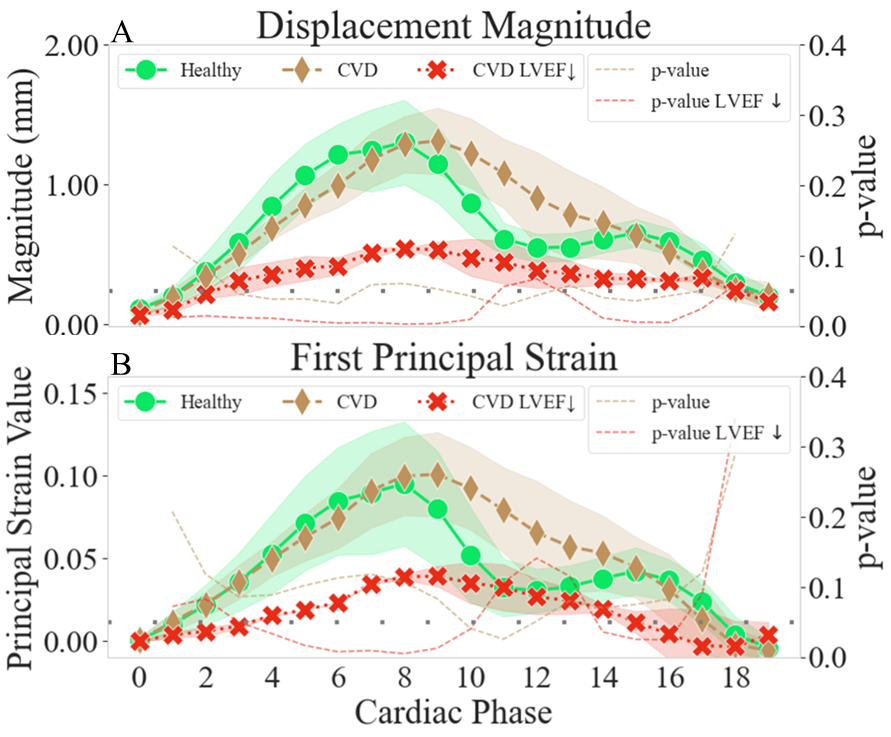}
  \caption{Comparison between healthy (green), CVD (brown), and CVD $\text{LVEF}_\downarrow$ (red) groups across the cardiac cycle and their respective t-test $p$-values between healthy/CVD (dotted brown) and healthy/CVD $\text{LVEF}_\downarrow$ (dotted red). Plot \textit{A} shows the DVF magnitude and plot \textit{B} the first Green-Lagrangian principal strain values. The values shown are after registering each case to the atlas to reduce variance from LA size differences.}
  \label{fig:la_mag_strains}
\end{figure}

\begin{table}[]
\centering
\caption{Average and standard deviation of image registration results for the LA contour mask across the cardiac cycle over three runs, along with the number of trainable parameters for each image registration model.}
\label{tab:registration_results}
\resizebox{1.0\columnwidth}{!}
{%
\begin{tabular}{@{}lccc@{}}
\toprule
\textbf{Model} & \textbf{Hausdorff Distance (mm)} & \textbf{Dice Score} & \textbf{\#Params} \\ \midrule
ANTs           & $2.57 \pm 1.37$                            & $0.72 \pm 0.12$           & 0           \\
Vxm            & $4.70 \pm 1.62$                            & $0.77 \pm 0.09$           & 2.8M        \\
Vxm-seg        & $4.23 \pm 1.86$                            & $0.72 \pm 0.12$           & 2.8M        \\
Aladdin-R      & $\mathbf{1.14 \pm 0.32}$                   & $\mathbf{0.98 \pm 0.02}$  & 2.2M        \\ \bottomrule
\end{tabular}%
}
\end{table}

\subsection{LA Deformation and Strains}
When spatially averaged across the LA, the time course of the LA principal strain values closely tracks that of LA volume (see SupFig \href{https://github.com/cgalaz01/aladdin_cmr_la/blob/main/supplements/README.md#SupFig2}{2}). The volume changes translate into clinically important global metrics such as the LAaEF and LAEF. Our 3D scans analysed with Aladdin lead to estimates of LAeEF and LAEF in good agreement with the traditional estimates from 2-chamber, 4-chamber, and 3D views - see Supplementary Table (SupTab) \href{https://github.com/cgalaz01/aladdin_cmr_la/blob/main/supplements/README.md#SupTab1}{1} and SupFig \href{https://github.com/cgalaz01/aladdin_cmr_la/blob/main/supplements/README.md#SupFig3}{3}. 3D LAaEF has the highest Pearson correlation with the 4-chamber estimates, achieving a value of 0.83. In contrast, the 3D LAEF shows high correlation with estimates from all long-axis views, with a value of 0.93, 0.87, and 0.98, respectively for the 2-chamber, 4-chamber, and 2/4-chamber. Additionally, we found that 3D LAaEF and LAEF correlate more strongly with LVEF than its 2D counterparts, with correlation values of 0.70 and 0.72, respectively.

The normalized mean principal strains measured using Aladdin and GLS estimates from the 2-chamber and 4-chamber views for the CVD cases are qualitatively similar in most cases - see SupFig \href{https://github.com/cgalaz01/aladdin_cmr_la/blob/main/supplements/README.md#SupFig4}{4}. These GLS estimates are consistent with those reported in the literature \cite{mualuaescu2022left,gan2018left,kim2020left}, also taking into consideration the effects of aging \cite{boyd2011atrial}. Four cases, which include both CVD $\text{LVEF}_\downarrow$ cases, show noticeable discrepancies, primarily during the conduit and booster-pump phases, which are likely to come from the incomplete LA coverage provided by the long-axis analyses. Furthermore, the peak LA strains at the reservoir and conduit phases from the 2-chamber and 4-chamber views may occur at different cardiac phases. This variability is likely due to differences in the scanning planes and the irregular, complex structure of the LA.

When analysing 3D strains in more detail, we observed that the atria of the healthy and CVD groups undergo different deformations across the cardiac cycle, especially in the conduit and active phases (Figure \ref{fig:la_mag_strains}). In the imaged cohorts, the atria of healthy subjects empty faster than that of the CVD patients leading to a more marked active contraction phase. This difference in deformation time course cannot be discerned using EF metrics. In contrast, CVD $\text{LVEF}_\downarrow$ cases showed a global decrease in LA function with low filling and slow emptying - see Figure \ref{fig:la_mag_strains}). 

Overall, the healthy cases had average DVF magnitudes of $0.67 \pm 0.40$ mm and principal strain values of $0.04 \pm 0.04$ across the whole cardiac cycle. The CVD cases had slightly higher averages of $0.71 \pm 0.43$ mm and $0.05 \pm 0.04$. Finally, the CVF $\text{LVEF}_\downarrow$ cases had much lower values of $0.34 \pm 0.15$ mm and $0.02 \pm 0.02$. The observed differences between the healthy and CVD cases are significant during the conduit phase (DVF: t-test $p < 0.05$; principal strain: t-test $p < 0.05$), while for CVD $\text{LVEF}_\downarrow$ cases, they are significant during the reservoir and booster-pump phases (DVF: t-test $p < 0.005$; principal strain: t-test $p < 0.05$).

Aladdin's main strength lies in its ability to provide otherwise unavailable insights into regional LA function and deformation, beyond global strain metrics. One of the findings from this feasibility study is that larger DVFs and principal strain values are observed at the anterior wall compared to other LA regions (see SupVid \href{https://github.com/cgalaz01/aladdin_cmr_la/blob/main/supplements/README.md#SupVid8}{8}). In contrast, the smallest DVF and principal strain values can usually be found on the LA roof and posterior wall (see SupVid \href{https://github.com/cgalaz01/aladdin_cmr_la/blob/main/supplements/README.md#SupVid8}{8}). The DVF and principal strain fields are spatially and temporally smooth, although the principal strain directions show a complex spatial alignment qualitatively consistent with LA myofibre orientations \cite{roney2019universal,pashakhanloo2016myofiber}.

%The calculated principal strain values are spatially and temporally smooth. Healthy cases (see representative case \ref{vid:sup_healthy_individual} second row) have higher first principal strain values on the anterior wall, consistent with the observed higher DVFs there. The smallest DVF and principal strain values can be usually found on the LA roof and posterior wall. In contrast, CVD cases (see representative case \ref{vid:sup_cvd_individual} second row) show more spatially homogeneous DVF and principal strain values. 

% strain directions
%The principal strain directions predominantly align to the circumferential direction across the cardiac cycle (see \ref{vid:sup_healthy_individual} and \ref{vid:sup_cvd_individual} second row). The alignment with the short axis plane is largest at the end of the reservoir phase.

%As with the DVF, we also observe differences in the average Green-Lagrangian principal strain value between the healthy ($0.09 \pm 0.05$) and CVD ($0.02 \pm 0.04$) groups (see Figure \ref{fig:la_mag_strains}B). These are significant during most of the reservoir phase (phases 0-6) and the booster-pump phase(15-18). The healthy cases have marked peak strains at the end of the reservoir phase (phases $\sim$7-9) and at the start of the booster-pump phase (phases $\sim$11-13). On the other hand, CVD cases lack a marked strain peak at the booster-pump phase and have overall smaller principal strain values.

\subsection{LA Atlas}
\begin{figure*}[!ht]
\centering
  %\textbf{Atlas}\par\medskip
  \includegraphics[width=0.9\textwidth]{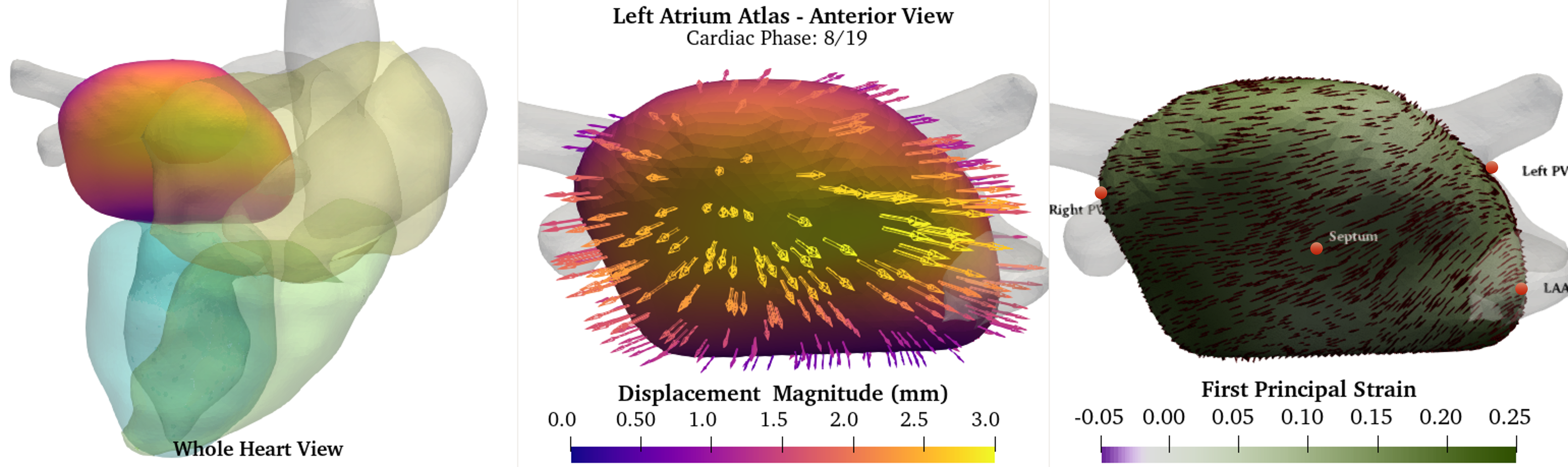}
  \caption{\textbf{Atlas of the LA Displacements and Principal Strains:} From left to right: 1) Whole heart view with the LA atlas, 2) the generated DVF, and 2) the estimated first principal strains for the anterior view at the end-reservoir phase (phase 8). The atlas across the cardiac cycle can be viewed in SupVid \href{https://github.com/cgalaz01/aladdin_cmr_la/blob/main/supplements/README.md\#SupVid9}{9}}
  \label{fig:la_atlas}
\end{figure*}

We were able to successfully register the LAs of all 10 healthy volunteers to a common space and calculate average DVF and strain biomarkers in this atlas. The resulting DVF and principal strain atlas can be seen in Figure \ref{fig:la_atlas} for the conduit phase and across the cardiac cycle in SupVid \href{https://github.com/cgalaz01/aladdin_cmr_la/blob/main/supplements/README.md#SupVid9}{9}.

We obtained in the atlas an average (across space and time) DVF magnitude of $0.67 \pm 0.39$ and principal strain value of $0.04 \pm 0.02$. The atlas fields preserve the observed properties in the DVFs and strains, showing temporal and spatial smoothness and higher displacements on the anterior wall and very low displacements in the roof. The three phases of the LA (reservoir, conduit and booster pump) remain clearly distinct (see SupVid \href{https://github.com/cgalaz01/aladdin_cmr_la/blob/main/supplements/README.md#SupVid9}{9}).

% The average across space and time principal strain value is $0.09 \pm 0.05$. We observe higher first principal strain values at regions of the anterior wall and at the end of the reservoir phase, as in the individual healthy LA cases (see representative case in \ref{vid:sup_healthy_individual}). The CV of the first principal strain is $1.30 \pm 4.84$ with larger variation at regions with small mean principal strain values that around found predominantly on the posterior wall and the start of the reservoir and end of the boost-pump phases. 

\begin{figure*}[!ht]
\centering
  %\textbf{CVD Regional Analysis}\par\medskip
  \includegraphics[width=0.9\textwidth]{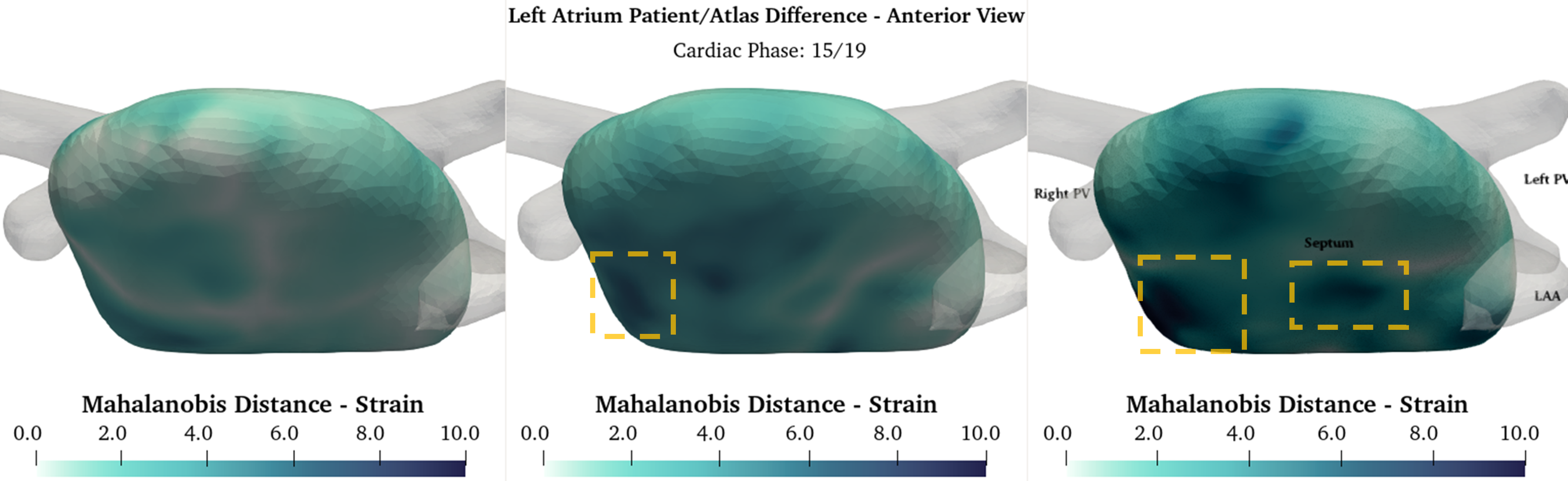}
  \caption{Mahalanobis distance of the first principal strain between the atlas distribution and the estimates for three representative patients with, from left to right: myocarditis, myocardial infarction (CVD $\text{LVEF}_\downarrow$) and non-ischaemic cardiomyopathy. The anterior view at the start of the boost-pump phase is shown. Regions with a high Mahalanobis distance are highlighted with yellow boxes, which may indicate areas of regional functional abnormality.}
  \label{fig:la_atlas_diff}
\end{figure*}

The atlas captures the normal values and ranges observed in the healthy cohort and allows regional assessments of individual cases. We assess this application of the atlas by identifying abnormal LA motion signatures on three patients with myocarditis, myocardial infarction (CVD $\text{LVEF}_\downarrow$ case) and non-ischaemic cardiomyopathy, respectively. The strain MD of these three cases can be seen in Figure \ref{fig:la_atlas_diff} at the start of the boost-pump phase. During the boost-pump phase, the LA is actively contracting to empty blood, which provides an informative snapshot of its functional characteristics. Displacement and strain MD across the cardiac cycle for three representative healthy volunteers can be seen in SupVid \href{https://github.com/cgalaz01/aladdin_cmr_la/blob/main/supplements/README.md#SupVid10}{10},
\href{https://github.com/cgalaz01/aladdin_cmr_la/blob/main/supplements/README.md#SupVid11}{11}, and
\href{https://github.com/cgalaz01/aladdin_cmr_la/blob/main/supplements/README.md#SupVid12}{12} and the three previously mentioned CVD cases can be seen in SupVid
\href{https://github.com/cgalaz01/aladdin_cmr_la/blob/main/supplements/README.md#SupVid13}{13},
\href{https://github.com/cgalaz01/aladdin_cmr_la/blob/main/supplements/README.md#SupVid14}{14} and
\href{https://github.com/cgalaz01/aladdin_cmr_la/blob/main/supplements/README.md#SupVid15}{15}, respectively. The regions with the most atypical deformation patterns correspond to the largest MD with respect to the healthy volunteer atlas. We observed that CVD cases generally have higher DVF MD and first principal strain MD on the anterior wall during the end of the reservoir and start of the booster-pump phases, compared to healthy cases. Additionally, CVD cases show higher MD values on the posterior wall (see SupVid \href{https://github.com/cgalaz01/aladdin_cmr_la/blob/main/supplements/README.md#SupVid15}{15}) across the cardiac cycle. However, it is important to note that regions with small or no deformations, such as the posterior wall, may exhibit unreliable MD values because the mean coefficient of variation is close to 0. While the locations of large MD often overlap in the DVF and principal strain value maps, the DVF MD typically covers a larger region. 

% This is caused by the fact that near-zero DVF values are interpreted the same by the resampling module, i.e. no deformation.

\section{Discussion}

% summary
In this study, we present a novel non-invasive framework to study LA regional and global motion and strains, based on high-resolution 3D Cine MR images and comprehensive image analysis workflow, \textit{Aladdin}. Aladdin consists of: a segmentation network, nnU-Net, to accurately identify the LA contours; an automatic image registration network to to extract LA DVFs, Aladdin-R; automatic strain tensor calculation from the DVFs; and, finally, the creation of a representative DVF and strain LA atlas. We show the feasibility of this approach by analysing images from 10 healthy volunteers and 8 patients with CVD and an indication for clinical MRI. Our technical feasibility results highlight the potential of this integrated methodology to provide clinicians and researchers with new regional biomarkers of LA motion and deformation.

%%%%%%

\subsection{Comparison with the Literature}
As full-coverage Cine MRI of the LA is a new and emerging research field, there is limited previous work using this modality to compare to Aladdin. Varela and colleagues \cite{varela2017novel} obtained the 3D displacements from high-resolution 3D Cine using the MITT \cite{mitt} on the LA segmentation maps. Morris et al, \cite{morris2020image}, on the other hand, focused on the identification of fibrotic tissue using full LA coverage MRI acquired using thick 2D slices. Displacements were estimated using ANTs \cite{avants2009advanced}. In the current study, Aladdin-R led to more accurate DVF estimations than these methods or other existing conventional or data-driven methods - see SupVid \href{https://github.com/cgalaz01/aladdin_cmr_la/blob/main/supplements/README.md#SupVid6}{6} and \href{https://github.com/cgalaz01/aladdin_cmr_la/blob/main/supplements/README.md#SupVid7}{7}. 

% comparison with literature (registration)
Aladdin's LA motion and deformation measurements are in good general agreement with global long-axis Cine estimates and the literature from 2D STE. Specifically, highest strain values are achieved during the end of the reservoir phase and a smaller peak during the start of the booster-pump phase \cite{nemes2019normal,bao2022left}. Additionally, the largest motion is observed on the anterior rather than the posterior side of the LA wall, and smallest motion on the roof \cite{kuklik2014quantitative}. The principal strain directions are predominantly parallel to the mitral valve plane throughout the cardiac cycle \cite{nemes2019normal,bao2022left}. We speculate whether Aladdin may additionally contribute to the estimation of LA myocardial fibre directions, which are important determinants of atrial electrophysiology and biomechanics. In particular, it would be interesting to compare our imaging-based estimation of the principal strain directions during the active contraction phase with the LA fibre orientation directions proposed using other methods \cite{roney2019universal,varela2013determination,krueger2011modeling,ho2002atrial}.

% \cite{voigt20192,plavsek2022agreement,satriano2017clinical}

% registration and segmentation
As we use a novel high-resolution imaging protocol, we have a comparatively small dataset of 18 subjects, unsuitable for traditional deep learning techniques. Online learning, employed in both the supervised nnU-Net and the weakly supervised Aladdin-R, is a natural choice to overcome issues of limited data and effectively address the large (compared to LV) structural and functional variability of the LA. As the networks are optimized on a subject-by-subject basis, they reduce biases from the training data and can continue to be employed in future studies to analyse images from subjects with various clinical presentations. The proposed methods are also well suited for future analyses of the motion and deformation of the right atrium or right ventricle (which are often also approximated as thin surfaces). 

In this study, we focus on DVFs and principal Green-Lagrangian strain values (with ventricular end-diastole as the Lagrangian reference frame) to characterize LA motion. Other related physical variables instead, such as strains in an Eulerian frame, the rate of change of the strain metrics \cite{hoit2014left}, or projections of these variables into a cylindrical reference frame as often calculated in the LV \cite{morales2021deepstrain}, can all be easily computed from the metrics we present in this paper. Future studies should identify which biophysical variables are most useful in different clinical applications. 

% atlas
The proposed atlas allows rapid and reproducible assessments of LA DVF and/or strain irregularities, as manual comparison of 3D DVFs and strains is not only labour-intensive but also prone to errors. In the current study, we use the MD to identify regions of the LA with abnormal biomarkers. This identification is likely to be more robust when several deformation metrics (e.g., principal strain value and DVF magnitude) are considered simultaneously, as will be explored in future studies. This analysis can be performed using the proposed atlas or other techniques for analysing regional LA information, such as the universal atrial coordinate 3D LA surface mapping system \cite{roney2019universal}.

%%%%%%

% atlas for fibrosis identification
Regional analyses of LA motion and deformation are likely to be clinically valuable for the identification of dense fibrosis or scar in the LA, as these regions are stiffer than healthy myocardium. Scar identification using LGE-MRI is generally reproducible, but quantifying diffuse myocardial fibrosis is highly subjective and less reproducible \cite{chubb2018reproducibility,fahmy2021improved}.

\subsection{Limitations and Future Work}

In future iterations of Aladdin, the segmentation and image registration tasks could be joined, as they are closely intertwined. This could improve overall performance as has been shown for the LV \cite{sinclair2022atlas}. Trained on image pairs to identify voxel displacement, Aladdin-R is agnostic to the cardiac phases provided and lacks temporal regularization. Although it already demonstrates strong temporal smoothness, its performance may be enhanced by tracking each voxel across all phases, similarly to \cite{wang2023tracking}. Lastly, DVF and strain estimates could be further improved by using biomechanics-informed regularization \cite{qin2020biomechanics}, perhaps in a physics-informed NN setting \cite{herrero2022ep,lopez2023warppinn}. All these refinements should contribute towards a more reliable DVF and strain estimation with little overheads to the processing time.

We plan to make a more representative atlas by acquiring and registering additional images of healthy volunteers with varied backgrounds and ages to update the atlas. The current atlas of LA motion uses images from 10 subjects, acquired in a single scanner, from a narrow age range and limited ethnicities that are not representative of the general population. This age gap (healthy: 30.4 $\pm$ 4.7 vs CVD: 53.8 $\pm$ 15.7) can also impact our current analysis, as older subjects tend to have reduced global LA conduit function \cite{sohns2020atrial}. 

We will also obtain more images of CVD cases, specifically AF cases. This will allow us to clinically evaluate the atlas' ability to identify diseased-induced functional abnormalities and further explore the links between atrial regional strain dysfunction and fibrosis. 

LGE-MRI is used for detecting fibrosis but is not widely used in the atria. Following further analyses, 3D Cine MRI could be an alternative or complementary technique to LGE-MRI to identify LA fibrosis. This could have important clinical applications in AF treatment stratification, the personalization of catheter ablations \cite{sohns2020atrial} and potentially early characterization of the AF risk phenotype \cite{shen2019atrial}.

As is typical of Cine MRI-based motion analysis, we are not tracking the displacement of material points and use image registration methods with regularization to determine the displacements of the LA wall. This is expected to have little impact on the motion analysis in most circumstances. It may however lead to inaccurate strain estimates in the presence of atrial devices, severe focal atrial pathology or in other circumstances in which the smoothness assumptions of the regularization techniques may be compromised. 

% A comparison between LA deformation metrics derived from Cine MRI and other modalities will be performed to determine 3D Cine MRI's clinical value. STE is the current gold-standard technique to assess LA motion. 3D STE enables full coverage scanning of the LA, but, when compared to 2D STE, suffers from lower spatial resolution, reduced image quality, and decreased reproducibility \cite{voigt20192,satriano2017clinical}. 

%This will provide deeper insights into the relationship between LA deformation metrics and the presence of fibrotic tissue, and identify a potential clinical application for Aladdin.

% Another limitation is the variability of the left and right PVs, and LAA \cite{ho2012left}, which functionally affect both the atlas construction and abnormality detection on an individual basis. By averaging the positions of these landmarks across cases, we may lose precision. Future research will explore advanced methods for landmark positioning to enhance the accuracy of the atlas.

\section{Conclusions}

We present a method to estimate 3D global and regional strains and displacements of the left atrium. Using high-resolution 3D CINE scans, we propose Aladdin, a tool to reliably provide 3D DVF and strain maps of the entire LA and map the information to a common atlas space. We expect this will allow the discovery of novel clinical prognostic and diagnostic biomarkers of LA function and improve our understanding of the atria's involvement in cardiovascular disease processes.

\appendices

\section*{Acknowledgment}
This work was supported by the UK Research and Innovation (UKRI) Centres of Doctoral Training (CDT) in Artificial Intelligence for Healthcare (AI4H) \url{http://ai4health.io} (Grant No. EP/S023283/1), the NIHR Imperial Biomedical Research Centre (BRC), and the British Heart Foundation Centre of Research Excellence at Imperial College London (RE/18/4/34215).

% \section*{References and Footnotes}

\bibliographystyle{IEEEtran}
\bibliography{_main}

\end{document}